%% file: paper.tex
\documentclass[]{fairmeta}

\input{template/package}
\usepackage{template/macro}

\definecolor{myboxcolor}{RGB}{230,245,255}
\title{Automatic Textbook Formalization}

\author[1,2]{Fabian Gloeckle}
\author[1,2]{Ahmad Rammal}
\author[1]{Charles Arnal}
\author[1]{Remi Munos}
\author[1]{Vivien Cabannes}
\author[1]{Gabriel Synnaeve}
\author[2,3]{Amaury Hayat}

\affiliation[1]{\small{FAIR, Meta}}
\affiliation[2]{\small{CERMICS, ENPC, Institut Polytechnique de Paris}}
\affiliation[3]{\small{Korea Institute for Advanced Study}}

\abstract{
We present a case study where an automatic AI system formalizes a textbook with more than 500~pages of graduate-level algebraic combinatorics to Lean.
The resulting formalization represents a new milestone in textbook formalization scale and proficiency, moving from early results in undergraduate topology and restructuring of existing library content to a full standalone formalization of a graduate textbook.
The formalization comprises 130K lines of code and 5900 Lean declarations and was conducted within one week by a total of 30K Claude 4.5 Opus agents collaborating in parallel on a shared code base via version control, simultaneously setting a record in multi-agent software engineering with usable results.
The inference cost matches or undercuts what we estimate as the salaries required for a team of human experts,
and we expect there is still the potential for large efficiencies to be made without the need for better models.
We make our code, the resulting Lean code base and a side-by-side blueprint website available open-source.  
}

\date{March 31, 2026}

\metadata[Code]{\url{https://github.com/facebookresearch/repoprover}}
\metadata[Formalization]{\url{https://github.com/facebookresearch/algebraic-combinatorics}}

\begin{document}

\maketitle
\section{Introduction}
\input{figures/churn_decls}

The \emph{replication crisis} of mathematics is the refereeing crisis. 
Under current incentive structures and time constraints, journal referees often cannot verify every proof in detail.
Indeed, the task of referees has been described not as checking the validity of all arguments, but as ``convincing themselves that the methods used in the paper are strong enough to prove the main results of the paper'' \citep{buzzard2020future}. 
While this system based on experts' intuition works remarkably well in practice and gaps can usually be filled in, the mathematical literature occasionally contains wrong proofs, sometimes even for wrong statements \citep[see the examples compiled in][]{voevodsky2014origins,buzzard2020future}.
Even important results by esteemed authors, applied and studied in seminars by large audiences, are not immune to overlooked mistakes. As Fields medalist Vladimir Voevodsky explained:
\emph{``A technical argument by a trusted author, which is hard to check and looks similar to arguments known to be correct, is hardly ever checked in detail.''}

\emph{Formalization} of mathematical results in proof assistant software is an ambitious long-term solution to this proof-checking problem.
While requiring significant investment in library fundamentals upfront, it reduces the trust layer to a small and tightly validated proof checker kernel,
and a few dozen lines of manually reviewed definitions and final theorem statements per publication.
However, due to the high levels of effort and expertise required, formal verification has so far been limited to a handful of landmark studies  \citep[e.g.][]{gonthier2005fourcolour,gonthier2013oddorder,gunther2022independence,vandoorn2023sphereeversion,commelin2023abstractionboundariesspecdriven,becker2025blueprintformalizationcarlesonstheorem}.

What is still missing for the aspiration of routine formalization of results in research mathematics?
First, massive amounts of foundational content from the textbook literature need to be made available in proof assistants.
Second, the cost of formalization -- in time, effort, and expertise -- must be reduced enough to make the endeavor worthwhile.
The development of Lean's mathematical library \emph{mathlib} \citep{The_mathlib_Community_2020}, sheds light on both challenges.
At roughly 2.2 million lines of code, it is the largest unified library of formalized mathematics, covering broad swaths of undergraduate material and selected topics at the graduate level. Yet, extending this coverage to a substantial fraction of the mathematical literature would require growth by several orders of magnitude.
Moreover, after an initial period of rapid acceleration, \emph{mathlib}’s expansion has stabilized to an approximately linear rate in recent years, questioning the scalability of a purely human-driven approach.

\emph{Automatic formalization} seeks to use artificial intelligence systems not only to accelerate mathematical research -- thereby increasing pressure on the largely manual refereeing system -- but also to formalize the output of mathematical research, freeing referees from the task of verifying proof correctness.
We can distinguish three generations of autoformalization systems.
\begin{enumerate}
\item \textbf{Bespoke models} for formalizing standalone mathematical statements in isolated environments. Such models have created the foundational machine learning datasets for formal mathematics \citep{xin2024deepseek,ying2024lean,dong2025stpselfplayllmtheorem} and have shown powerful in solving and formalizing IMO and Putnam competition problems \citep{lin2025goedel,chen2025seedprover15masteringundergraduatelevel,ren2025deepseekproverv2advancingformalmathematical,hubert2025olympiad,achim2025aristotleimolevelautomatedtheorem},
but lack the repository-level engineering skills needed to be directly applicable to textbook formalizations. 
\item \textbf{Agentic coding assistants} \citep{schick2023toolformer,yang2024sweagentagentcomputerinterfacesenable} powered by generalist models and deployed in a single-agent setup \citep{urban2026130klinesformaltopology}, possibly complemented by tools specific to the proof assistant \citep{liu2026numina}. Following recent progress in language model post-training, generalist frontier models are proficient at theorem proving in many formal systems, while their tool-use training allows them to interact with the proof checker in a straightforward way.
\item \textbf{Multi-agent scaffolds}
with several agentic coding models collaborating on a shared code base. So far, published explorations remain limited in scope (e.g., requiring human review of all theorem statements) \citep{wang2026m2fautomatedformalizationmathematical}, while closed-source efforts do not disclose sufficient details \citep{skuse2026watershed,chen2026felsconjecturesyzygiesnumerical}.
\end{enumerate}

Multi-agent scaffolds require a solution to the \emph{coordination problem}: How can a large swarm of agents be organized to make consistent progress on a shared project? Initial explorations in the realm of software engineering have produced remarkably large code bases, but also shown the coherence issues that arise when agents are insufficiently orchestrated \citep{Lin2026ScalingAgents,Carlini2026BuildingCCompiler}.

In this study, we propose a simple multi-agent scaffold that largely resolves these limitations by relying on battle-tested standard practices inherited from human collaborative software engineering:
\begin{enumerate}[label=\Alph*)]
\item Large-scale parallelization via sub-agents with well-defined task assignments.
\item Version control using \texttt{git} with trunk-based development on short-lived feature branches.
\item Pull request review by independent reviewers with clear code quality guidelines.
\item A merge queue with staging branches tested before merge, so that \emph{main always builds}.
\item An escalation and communication mechanism using a file-system-based \emph{issue tracker}.
\end{enumerate}
Shared files provide a direct, simple and universal communication protocol for agents without the need for bespoke solutions for message passing.
The use of source control provides version history and conflict tracking in shared code bases with an interface that agents naturally master.
Pull request review cycles appear to adequately address code quality issues and preventively reduce coordination problems at later stages.
Overall, our system largely resolves the  coordination problem, allows \emph{fearless concurrency} and enables effective parallelization at scale without the need for heavy orchestration logic in the Python code.

To give examples for the efficacy of the communication mechanism via an issue tracker, consider the following examples of issues encountered and resolved over the course of our case study, each of which pertaining to larger sections of code spanning mutiple definitions, lemmas or theorems.
\begin{itemize}
\item Replace a proof strategy by one that requires fewer hypotheses.
\item Refactor files to allow for code reuse and reduce duplication.
\item Restructure a proof to circumvent the creation of false sub-goals.
\item Replace a placeholder definition by the true definition, and refactor downstream proofs accordingly.
\item Transfer proofs related to a function operating on certain objects to a finer-grained one operating on sub-objects thereof.
\end{itemize}

Tested in a large-scale case study, our multi-agent workflow successfully verified a complete graduate-level textbook in algebraic combinatorics \citep{grinberg2025introduction} within a week of runtime, dwarfing human verification times on projects of comparable scale and underlining the effectiveness of our parallel setup.
At a cost of around \$100K, or around \$200 per page and \$300 per target theorem or definition, automatic formalization is increasingly becoming an economically viable option, not accounting for the likely efficiency improvements we expect in the near future.

\section{Method}
\label{sec:method}
\input{figures/summary}

\subsection{Proof engineering as an agentic task}
Interactive theorem provers such as Lean cast proof formalization as a (mathematical) engineering task.
Indeed, from a software engineering perspective, a formalization project is nothing more than a code repository in a special-purpose programming language that needs to build without errors (and whose target statements need to be verified).
Agentic coding LLMs are naturally optimized for this task: with tools for reading and editing files and running terminal commands, coding agents now routinely explore existing projects, make themselves familiar with the context and current status of the task, develop and debug code and add documentation on what was achieved and the next steps.
Adding the increasing familiarity of frontier models with interactive theorem provers, agentic coding LLMs become a natural choice for automatic formalization projects as demonstrated by \citet{urban2026130klinesformaltopology} in a formalization of parts of a topology textbook using Claude Code. The formalization used a single instance of a command line coding agent and covered the initial chapters and first nontrivial theorems of point-set topology as presented in \citet{munkres_topology_2018}.
The focus of this work is to \emph{scale up} this process: using hundreds of such formalization agents, we attempt to formalize longer, more advanced sources in the more challenging context of Lean's comprehensive mathematical library.

\subsection{Large-scale Collaboration in Formalization}
Large-scale formalization efforts have always been collaborative, from early major results
\citep[e.g.][]{gonthier2013oddorder,hales2017kepler}
to the development of Lean's \emph{mathlib} \citep{The_mathlib_Community_2020} and more recent Lean projects
\citep[e.g.][]{vandoorn2023sphereeversion,
commelin2023abstractionboundariesspecdriven,becker2025blueprintformalizationcarlesonstheorem}. The simple reason for this is that a single formalizer would need prohibitively long to accomplish such major efforts alone. The same holds true for LLM agents: for real-world latency constraints, single-agent formalization like conducted in \citep{urban2026130klinesformaltopology} does not scale and a multi-agent approach is needed. In the remainder of this section, we will explain the orchestration layer we used in this experiment.

\subsection{Sketching and Decomposition}
In initial experiments, we explored formalization using a decomposition approach as suggested in various works on LLM-based formalization \citep[e.g.][]{jiang2023draftsketchproveguiding,ren2025deepseekproverv2advancingformalmathematical}. Concretely, we experimented with a recent preprint on $Q$-circulant matrices \citep{li2026extension}. We instructed an agentic LLM to create the scaffolding for a full formalization consisting of all definitions and theorem statements.
We had it create a prompt for prover subagents and launch and supervise one for each theorem, merging proofs into the file until the formalization completed. This approach could easily be stabilized into a small code base that handles the decomposition, file buffer state and LLM API calls, but proved unable to scale beyond simple preprints or book chapters because of its limited context handling.

\subsection{Multi-Agent Development}
\label{sec:dev-agents}
To scale beyond single file formalizations, we therefore opted for a repository level approach just like human formalization projects have done. Specifically, we use \texttt{git} for version control and adopt a trunk-based development model where agents work on short-lived feature branches. The features are supposed to be as small and self-contained as possible to enable fast reviewing by reviewer agents and integration into the main branch.
We therefore settle on the following set of roles for feature development:
\begin{itemize}
    \item \textbf{Sketcher agents} receive a chunk (say, a chapter) of the source material and produce the formalizations of all definitions and theorem statements contained in it, omitting proofs with the \texttt{sorry} keyword. 
    \item \textbf{Prover agents} receive theorems with missing proofs and fill the proofs if possible. They can escalate problems with the statement formalizations, the ordering of statements in the file or missing helpers on the issue tracker, and also decide to fix such problems directly instead.
    \item \textbf{Reviewer agents} receive pull requests by sketcher and prover agents and decide whether they meet quality criteria. Concretely, we split the reviewing into two independent parts: a \emph{mathematical review} that focuses on whether the statements accurately reflect the statements from the source material, and an \emph{engineering review} that focuses in the practical aspects of code quality (conventions, naming, documentation, API design). A pull request is approved if it can merge without conflicts into the main branch, the repository builds on the branch and both reviews are approved, following standard practice in continuous integration workflows.
\end{itemize}
Reviewers can choose between the following \emph{verdicts}: approval for code that can be merged; requesting changes for code with minor issues that should be addressed; rejection for code that is beyond hope (flawed proof strategy, code lost during merge conflict handling, duplication of existing results). If any reviewer rejects, the pull request is suppressed. If any reviewer requires changes, the merge into main or the build fails, the pull request is returned to the agent with a description of the issue for fixing.

Agents work on short-lived \texttt{git worktree}s with the file state of their individual branches. By sharing the \emph{mathlib} installation, each worktree is comparably lightweight, containing only the file contents and build artifacts of the repository.

\subsection{Multi-Agent Coordination}
\label{sec:coord}
For coordination, observability and interpretability, we also implement a lightweight \emph{issue tracker} system: The repository contains a folder with short text files describing to-do items (issues). Agents are encouraged to tick the issues they resolve in their pull requests as resolved, and reviewers check on these judgments. We incorporate three other agent roles that play a minor role in the overall system but that we found helpful for observing the system at work:
\begin{itemize}
    \item \textbf{Maintainer agents} receive an open issue from the issue list and are tasked to ``make progress'' on the issue, either by resolving it fully or by committing helpful code that will make resolution easier down the line.
    \item \textbf{Triage agents} review open issues to identify those that have already been resolved and mark them accordingly.
    \item \textbf{Scan agents} scan the code base for ``global'' issues that should be addressed and add them to the issue tracker: e.g. code duplication, potential for refactoring, unmatched conventions, missing lemmas that would make the library easier to use.
    \item \textbf{Progress agents} likewise scan the code base but with the explicit goal of tracking progress on the target theorems specified in the beginning in order to prevent ``derailing'' and focus on side issues.
\end{itemize}
There is no inherent difference between pull requests that touch the Lean code and pull requests that touch the issue tracker folder (both can be modified in a single pull request), and all pull requests including those by maintainer, triage and scan agents are reviewed by the two reviewer agents types described above.

In each chapter of the source material, we identify important \emph{target theorems and definitions}, omitting helper lemmas that only exist for the purposes of a proof by means of a prompted LLM. Each chapter's target theorems are added to the issue tracker as proving tasks.

\subsection{Tool Use}
Agents have access to tools similar to the ones used by command line interface coding agents. The tools fall into four different categories:
\begin{itemize}
    \item \textbf{File tools:} listing files, reading chunks of files based on line number ranges, writing new files, editing files based on search and replace, editing files based on line number ranges, editing files with copy-pasting and cut-pasting based on line number ranges.  
    \item \textbf{Lean tools:} Lean snippet execution without the need for temporary files via the Lean REPL \citep{morrison_repl}, specialized search tools for Lean's \emph{mathlib}: mathlib file search, mathlib grep, mathlib declaration search with specialized regular expressions.
    \item \textbf{Git tools:} running a subset of \texttt{git} commands that make sure agents only have write access to their own branch (\texttt{git status, add, log, diff, commit, rebase, reset}, checking out a file from a \texttt{git} reference, showing conflicts.
    \item \textbf{Bash tool:} an allowlist of shell commands for comparably safe execution: \texttt{lake} for Lean builds, text processing commands such as \texttt{cat, head, tail, wc, grep, sort, uniq, cut, awk, sed, diff}, file system commands such as \texttt{ls, tree, basename, dirname}, utilities such as \texttt{date, sleep}. Commands can be piped but output cannot be written and rudimentary checks ensure the current working directory is not left.
    \item \textbf{Issue tracker tools:} two helper tools for the issue tracker: \texttt{create-issue} for creating a new issue file with a fresh UUID, and \texttt{list-issues} for showing a summary of all open issues.
\end{itemize}
In the design of all tools, we take care to add automatic truncation to prevent context overflow from a single misguided tool use.

\subsection{Technical Details and Issues}
\label{sec:tech-issues}
We relaunched the system several times due to the following performance-related issues. We use eight machines to distribute build process load and a shared network file system (NFS) directory to conveniently sync \texttt{git} status between them. This makes NFS bandwidth a likely bottleneck that sometimes made \texttt{git worktree} copies and \texttt{lake build} processes time out and fail, in particular when runs were relaunched. For this reason, we needed to add delays and jitter to prevent thundering herd effects upon resumption. 

Moreover, the single-threaded shared merge queue is a choke point for the system: pull requests need to queue for a long time, and when they are ready to be merged, they may no longer be compatible with the current main branch and need to be requeued after rebasing and fixing. This is a well-known issue with branch-only testing in continuous integration that is more thoroughly solved by automatic staging branches with automatic PR batching and failure bisection, such as by \texttt{bors bot} used in the \emph{mathlib} repository \citep{chen2025growing}.

\section{Case Study: Algebraic Combinatorics Textbook Formalization}
We study the efficacy of multi-agent proof engineering systems by attempting a full formalization of \citet{grinberg2025introduction}, a graduate textbook on algebraic combinatorics of more than 500 pages. We chose this source because it is fully accessible from the machinery provided in Lean's \emph{mathlib} but to very large parts not contained in \emph{mathlib}. It also has complete proofs with few external dependencies and, thanks to the author, is in the the public domain. While less flashy than a formalization of a recent research result, we emphasize that instead of more landmark projects, routine formalization of research mathematics first requires routine formalization of foundational texts at a faster rate than human \emph{mathlib} contributors can provide. Working on well-chosen sources which are directly accessible from \emph{mathlib} is not a simplification, but the principal \emph{modus operandi} of literature autoformalization systems of the future.

\subsection{Avoiding Contamination}
It is vital to evaluate coding agents on previously unsolved tasks. While training data contamination is already a concern, an even more direct form of ``cheating'' is direct use or copying of code from repositories and libraries accessible to the agent at test time. For instance, an exploratory project on a multi-agent system working on a web browser \citep{Lin2026ScalingAgents} has been accused of heavily relying on open-source projects that do all the heavy lifting \citep{Gerard2026CursorLies}. Likewise, a project on an LLM-written C compiler \citep{Carlini2026BuildingCCompiler} has been criticized for relying on GCC for the 16-bit part of its claimed bootable Linux image creation \citep{VaughanNichols2026AnthropicCCompilerOpinion}. In concurrent work, \citet{wang2026m2fautomatedformalizationmathematical} propose a textbook autoformalization system which produces formalizations of real and complex analysis textbooks relying on \emph{mathlib}, which already extensively covers both topics.

In this work, we choose a textbook that appears to be largely disjoint from content contained in \emph{mathlib} to avoid contamination and cheating. Where there is overlap, we do encourage the system to reuse existing theories, resulting in a few superficial files that essentially restate theory about formal power series or the fundamental theorem of symmetric functions from \emph{mathlib}.
Our agents do not have internet access, which further reduces the possibility of cheating.

\subsection{Overview of the Formalization Challenge}
To track progress of the formalization effort and not mistake activity for achievement, we designate 340 definitions and theorems in the Latex source as \emph{targets} of the formalization. For textbook formalization projects where there is not a single ``main theorem'', we believe this approach to be an accurate measure of coverage while at the same time allowing enough flexibility in the precise proof architecture and reducing review workload compared to treating every statement as a target.

\input{figures/contents}

Figure~\ref{fig:contents} gives a short overview of topics covered in the source textbook and its formalization. The full contents of the formalization can be found in the source textbook, the Lean source code, the Lean documentation and an auto-generated blueprint website.

\subsection{Results}
\input{figures/loc_hists}
\input{figures/agent_tokens}

The formalization concluded successfully after one week of runtime, formalizing all 340 target theorems and definitions.
Four initial targets were correctly reclassified as exercises in the source and not attempted -- we focus exclusively on formalization, not on problem solving.
As shown in Figure~\ref{fig:churn-decls}, both the overall number of lines and of declarations grow at an approximately constant rate, reaching 130K lines and 5,900 declarations at the end of the experiment. 
Table~\ref{tab:stats} contains additional statistics on the volume of the resulting formalization.

Our strategy of breaking down tasks into easily reviewable segments proved successful. As illustrated by the line count histograms in Figure~\ref{fig:hists}, even the largest agent-submitted pull requests did not exceed a few hundred lines of code.

This experiment used Claude 4.5 Opus, an agentic coding model capable in Lean, but we expect current frontier models to behave roughly comparably on this task.

We generally observed a high statement formalization quality bar imposed by the review agents and manually spot-checked key statements and definitions. This gives us a high degree of confidence in the overall correctness of the formalization, but it remains a possibility that individual theorems and definitions have not been translated semantically faithfully. We release the formalization repository alongside an interactive formalization blueprint website that will allow the research community to compare statements and definitions side-by-side. 

\subsection{Agent Outcomes}
\label{sec:outcomes}
We classify the possible outcomes of an agent run as follows: 
Overall success means that a pull requests is filed, approved and merged (\emph{Merged}).
In the other cases, agents can time out if they exceed a limit on the number of conversation turns (128--512, \emph{Max Iterations}) or on the number of pull request review cycles (10, \emph{Max Revisions}).
If agents return without a pull request \emph{(No PR)}, they can indicate that their work is blocked by issues that have to resolved first \emph{(No PR (blocked))}.
For technical reasons, the merge process may fail (Section~\ref{sec:tech-issues}) or agents may be dropped between restarts \emph{(Aborted)}.
This can also happen after a pull request has already been approved \emph{(Approved)}.

\subsection{Cost}
\input{figures/outcome_tokens}

\input{figures/outcomes_runtime}
We break down the total amount of compute used in the experiment by the number of agents per type (Figure~\ref{fig:agents-by-type}),
the total amount of input and output tokens per agent type (Table~\ref{tab:agent-tokens}),
the total number of tokens per agent outcome (Table~\ref{tab:outcome-tokens}) and its evolution over the experiment's runtime (Figure~\ref{fig:outcomes-runtime}).

The full formalization required a total of 83B input tokens (with multiple counting in multi-turn dialogs) and 561M output tokens. Without token caching, this would correspond to a total cost of \$430K. While our logs unfortunately do not contain token caching statistics, a back-of-the-envelope calculation (App.~\ref{app:cache}) suggests an approximate price of \$100K overall, \$14K out of which for output tokens in both cases.

This price tag approaches, and may even undercut, the one required for a formalization by a team of human experts. However, it remains prohibitive  to directly scale textbook formalization by orders of magnitude.

That said, this initial exploration likely overestimates the true cost of conducting the formalization by a large margin, for the following reasons:
\begin{enumerate}
\item \textbf{Iterative development:} In this exploratory case study, we developed and fixed the multi-agent orchestration code partially during the run, reacting to performance inefficiencies as they arose. The final code resolves several issues related to interruption and resumption, accidental work duplication, erroneous tool definitions, timeouts and thrashing that wasted compute during initial phases of the experiment. In particular, around half of the input tokens are spent on \emph{aborted} agents, i.e. agents that were not continued after restarts which happened during initial problems with timeouts and thrashing.
\item \textbf{Unnecessary work:} During later phases of the experiment, agents would accidentally attempt proofs for exercises and cited theorems, despite instructions to the contrary. This could be fixed with improved prompts, but significant numbers of agents worked on ultimately unnecessary parts of the code base. 
\item \textbf{Lack of dependency tracking:} In our experiment, agents appeared well aware of the dependency relations between proof tasks and other maintenance issues in the code base, frequently adding such remarks to issue descriptions and source code comments without having been prompted to do so. Our current code does not track and leverage this information, simply requeuing agents that reported their tasks to be blocked at the current state of the code base. A directed acyclic graph of task dependencies that determines agent launch order would improve inference efficiency, and could be kept in a consistent state using simple algorithmic checks and dedicated agents for fixing accidental dependency cycles.
\end{enumerate}
With the bug fixes integrated over the course of the experiment, prompts that more effectively prevent unnecessary work on exercises and cited theorems from the outset, a more conservative tuning of the degree of parallelism and the relaunch frequency for failed and blocked agents and further general improvements in the orchestration layer, we believe that a 3-10x decrease in inference cost can be achievable without the need for better models.

\subsection{Analysis: Provers vs Maintainers}
\input{figures/prover_maintainer}

Our multi-agent orchestration makes use of two complementary mechanisms for organizing task assignments. Initially, a simple setup relied exclusively on assigning unproved theorems to \emph{prover} agents with the task of filling in full proofs for the theorem statements previously formalized by \emph{sketcher} agents.
Later, we observed the need for a more flexible mechanism with which agents could exactly flag the blockers and issues they encountered. These would be represented as files in an issue tracker directory, pull requests to which are examined by math and engineering review agents just like the Lean code base. \emph{Maintainer} agents would then be assigned issues to resolve which can represent more global tasks and more fine-grained tasks than the ones assigned to \emph{prover} agents (e.g. refactorings or filling individual \texttt{sorry} gaps in a partial proof).

It is an interesting question whether \emph{prover} agents are required at all, or whether their role should be subsumed by \emph{maintainer} agents in subsequent iterations. While we have no definite answer to this question, we point out their differences in Figure~\ref{fig:prover-maintainer}. \emph{Prover} agents exclusively work on single Lean files by design, while \emph{maintainer} agents can carry out heavy refactorings involving up to 14 files. Conversely, \emph{maintainer} agents are more frequently involved in ``bookkeeping pull requests'' that exclusively touch coordination files (for better or worse). Note that the cluster of \emph{prover} agents that do not touch any coordination files is related to the late addition of the feature in the experiment.

\section{Observations and Learnings}
\label{sec:learnings}
In collaborative software engineering, there is a tension between \emph{agility}, i.e. local agency and proactivity, and \emph{coherence}, i.e. shared structure and consistency in the overall codebase \citep[e.g., among many others][]{brooks1995mythical,boehm2003balancing,skelton2019team}. A multi-agent mathematical formalization system faces the same trade-offs: tasks can easily be broken down by proof targets, but global coherence needs to be maintained when it comes to shared conventions, preventing code duplication or interoperability of sub-theories with downstream use cases. We illustrate observed failure modes from missing coordination and too aggressive parallelization, both on the level of mathematical semantics as well as of process orchestration.

\subsection{Coherence Failures in Mathematical Formalization}
In mathematical formalization projects, definitions are often deemed the parts that are hardest to write and most depended upon with the project. Besides initial failures where agents provided placeholder definitions (such as the identity function for a complicated map), we encountered the following two large-scale definition-related coherence issues over the course of the project.
\begin{itemize}
\item An $N$-partition data type was defined three times independently in the chapters on Schur polynomials, symmetric polynomials and the Pieri rules. All definitions were mathematically equivalent and provided similar local helper lemma APIs. After the issue was discovered, a shared definition was factored out into a separate file, temporarily increasing the count to four copies\footnote{evoking \url{https://xkcd.com/927/}}. Two files eventually migrated to the shared definition while for the third one, the migration was deemed to cumbersome due to the extended private API spanning 8k lines, and an equivalence was proved instead to bridge between the two versions. The refactoring of these remaining two definitions is tracked as an open issue. 
\item The most challenging part of the project was the formalization of Bender-Knuth involutions. Unlike the trivial case of $N$-partitions, the Bender-Knuth involution is hard to define formally. Indeed, the source requires five pages to define the map and prove that it maps to semi-standard tableaux again. The definition and codomain proof contain two concrete worked out examples and two additional drawings explaining the algorithm. Moreover, combinatorial algorithms are often explained very informally in mathematical text, e.g. next to a more formal treatment, the source also contains the following summarizing explanation:
\begin{quote}
Thus, $\beta_k(T)$ is obtained from $T$ by ``flipping the imbalance between free $k$'s and free $(k+1)$'s'' in each row of $T$ (so that a row that was heavy on free $k$'s becomes equally heavy on free $(k+1)$'s, and vice versa).
\end{quote}
Again, two separate definitions were given originally, one for the proof that skew Schur polynomials are symmetric and one for the Stembridge lemma. In this case, however, the definitions did not turn out strictly equivalent: one operates on full rows, one can be restricted to an initial prefix of columns. Moreover, while the prefix version took a wrong shortcut (ignoring matching conditions completely and flipping all allowable entries), the full-row version did implement matching but got the direction wrong for $k+1$ entries (selecting the rightmost excess $k+1$'s instead of the leftmost).

This inconsistent situation led to \emph{agent churn}: 
Where lemmas were provable with a broken definition, prover agents happily provided the proofs. Where the lemma was false for the broken definition, prover agents left comments, created issues and gave counter examples but did not investigate globally and were unable to give a clear path forward for fixing the underlying issue. This may have been exacerbated by ``myopic'' prompts focusing on proof progress and review agents insisting on preventing regressions in existing code, even though, in principle, a system had been put in place to flag underlying issues with the broader formalization context.
\end{itemize}

\subsection{Process-Level Coherence Failures}
Coherence issues cannot just appear at the mathematical semantic level. Missing organization can also lead to desynchronized work 

Specifically, \emph{rabbit holes} turned out to be a major issue where agents would -- against explicit order -- embark on proofs for cited theorems or unproved statements mentioned in passing remarks. Once a statement was formalized and not marked as \texttt{cited}, the proof gaps would be picked up by prover agents, issues would be created and the autoformalization machinery would be launched on an \emph{proof search task}, often considerably more difficult than the intended proof translation task. For example, this happened for Kasteleyn's formula for domino tilings where the agents started developing in broad and very incomplete strokes the theory of Pfaffians and the FKT algorithm, drawing resources away from actual proof targets.

Likewise, the single merge queue without merge batching turned into a bottleneck during phases of the run with high degree of parallelism. Agents would repeatedly be asked to update their code due to merge conflicts, giving up eventually and creating additional issues on the tracker. This would in turn spawn additional maintainer agents, further increasing the resource contention.

\subsection{Steering Project Coherence}
The \emph{agility-coherence trade-off} can be steered by more or less rigid prompts, roles and orchestration layers: When the orchestration is tight (provers and maintainers are required to document clearly in the issue tracker where they are blocked and what next steps should be), we observe diffusion of responsibility for overall progress and a degradation to ``LLM bureaucrats'' that use said mechanisms to document why their work is blocked by a distant issue. When orchestration is more flexible, we see more proactive agency, but at the cost of overall structure, code duplication and convention mismatches.

We observed our experiment live and to a certain degree adjusted and finetuned the method based on the issues encountered above a handful of times over the course of the run. In particular, we adjusted prompts,  added agent types, clarified roles and review policies and tuned orchestration hyperparameters.

As a general rule, according to our experience, more freedom can be left at the beginning of a formalization project in order to leverage parallelism, while the ``screws can be tightened'' when the release nears in order to clean up and harmonize the project. In other words, we perform ``simulated annealing'' on the formalization project. 

Concretely, the \emph{scan, triage} and \emph{progress} agents we implemented in order to provide this global guidance could only partially fulfill their functions: on the more global issues such as the ones described above, their high-level task descriptions (along the lines of ``check the repository and find issues / blockers'') prevented them from going as in-depth as needed to flag deeper mathematical and engineering concerns. We manually interrupted the run a few times and launched CLI agents with the task of producing an in-depth report on the status of target theorems, \texttt{sorry} occurrences in the files and their downstream dependencies, global proof strategy and individual analyses for each remaining target on its blockers, sometimes manually nudging to investigate further when smells were detected. We had these \emph{status agents} create new issues and a summary file that would then be the starting point for the multi-agent system upon resumption. While we did not implement status agents in the code base, we note that this should be trivial to add and our human input was negligible (we purposefully did not study the proof or its structure and only interact on a high level by prioritizing observations made by the agents and asking for details and extended analysis).

\section{Related Work}

Early works on automatic formalization explored how language models could be used to translate statements \citep{wu2022autoformalization,wang2025aria}, informal proofs written by other models \citep{jiang2022draft} or informal chains-of-thought written by the same model \citep{xin2024deepseek,wang2025kiminaproverpreviewlargeformal}. 

Recently, large-scale automatic formalization has received considerable attention in the literature with several works published over the course of weeks.
\citet{urban2026130klinesformaltopology} considers a single-agent setup that manages to formalize the first non-trivial parts of an undergraduate point-set topology textbook.
\citet{wang2026m2fautomatedformalizationmathematical} presents a closed-source, rigid two-phase multi-agent scaffold that first translates all statements and then fills in proofs in parallel. The case study uses selected chapters from introductory real analysis and convex analysis textbooks, each with significant overlap with the existing \emph{mathlib} library that is used in the formalizations.
\citet{mathinc2025gauss} formalize the proof of the strong prime number theorem with their \emph{Gauss} system and finish the formalization of the proof for optimal sphere packing in dimensions 8 and 24 after the project was set up and initially developed by a team of human experts. The proprietary system converted 40 pages of human research articles into 200K lines of Lean code using undisclosed amounts of compute \citep{mathinc2026sphere}.
In our work, in contrast, we study optimal multi-agent orchestration with an open-source system that has been stress-tested on more than 500 pages of graduate mathematics, completing a full book that to the largest part is disjoint from Lean's \emph{mathlib}.

\section{Discussion}

\subsection{Scaling Up?}
To realize the vision of routine formalization in research mathematics, the effort required per new publication must drop to a near-constant cost per page, rather than scaling with the size of its unformalized dependency tree.
Therefore, a critical prerequisite is the systematic formalization of shared mathematical infrastructure.
By establishing this groundwork beforehand, we can isolate the effort of formalizing novel research from the burden of reconstructing its foundations.
We estimate this shared corpus spans 1,000 to 10,000 textbooks, encompassing the literature necessary to define core objects, provide the tools to manipulate them, and allow formalizers to \emph{state} the vast majority of modern research theorems.

We believe that automatic formalization efforts should prioritize the underlying textbook corpus, a focus best understood by contrasting it with two prevalent alternatives: competition mathematics and novel research. While the past year has seen intense focus on competition mathematics \citep{chen2025seedproverdeepbroadreasoning,hubert2025olympiad,achim2025aristotleimolevelautomatedtheorem}, such problems typically present highly specialized, self-contained puzzles. Moreover, while formalizing cutting-edge research is the ultimate goal, doing so directly often yields narrow, ad hoc definitions tailored to specific theorems. 
Unlike these isolated, single-shot challenges, textbook formalization represents an investment into shared infrastructure.
Textbooks are explicitly designed to maximize generality and reusability, meaning their formalization generates compounding, reusable mathematical knowledge that continues to pay off long-term, as demonstrated by Lean's \emph{mathlib}.

Scaling up automatic formalization to thousands of textbooks presents challenges beyond sheer size:
textbooks need to be selected and ordered in a rough dependency order, mismatched conventions need to be spotted and bridged, definitions and theorems need to be formulated at the right level of generality.
While different in scale, these questions are very similar in kind to the issues encountered in this case study at the level of a single textbook (Section~\ref{sec:learnings}) and we are hopeful that the coordination mechanisms presented in this work will continue to prove useful across scales (Section~\ref{sec:method}). Indeed, we have observed the issue tracker to organize the multi-agent system from the local level of single proofs to cross-chapter refactorings, involving changes to definitions and theorems used in several files and moving, unifying and deduplicating content across the entire repository.

A formalization of the core infrastructure of modern mathematics is invaluable -- not just for human mathematicians and verified research publications, but critically for the advancement of machine learning. First, an extensive formal library provides high-quality training data, which has frequently been shown to improve reasoning capabilities across the board \citep{zhang2024unveilingimpactcodingdata,ma2023trainingstagedoescode,yang2025codethinkthinkcode}. Furthermore, replacing heuristic final-answer comparisons with robust formal verification in reinforcement learning runs offers the most rigorous feedback signal possible for proof correctness. Ultimately, scaling inference to unprecedented magnitudes without the risk of undetected logical errors paves the way for automated research and discovery in the mathematical sciences. Gains in formalization capabilities directly translate to larger formal datasets, creating a virtuous cycle of self-improvement that will continue as long as unformalized mathematical results remain.

\subsection{Conclusion and Outlook}

Formalization is an extremely promising solution to the proof verification bottleneck that is increasingly plaguing mathematical research.
It gives correctness guarantees for proofs and reduces the review surface to definitions and theorem statements only, which often represent a tiny fraction of the overall material contained in a research article.
On top of this, formalization enables new ways of conducting mathematical research at scale:
It enables massive collaboration \citep{bolan2025equational} as well as
semantically aware algorithms that generate, search, refactor and transform proofs in ways that would be tedious or impossible to carry out by hand.

However, for the large-scale formalization of research mathematics to become a viable option, first large quantities of textbook content need to be formalized.
In this case study, we examined the formalization performance of contemporary language models for agentic coding in a highly parallel multi-agent setup.
We chose a graduate textbook as a source that is directly accessible from Lean's \emph{mathlib} but whose results are to large parts not yet contained therein.
The parallel formalization using Claude 4.5 Opus models exceeded our expectations: all 340 target definitions and statements have been formalized and proved.
At a cost of \$100K, the formalization 
already matches or undercuts the price required to pay a team of human experts and it required one week, dwarfing typical human project timelines.
Moreover, we identified and improved several performance issues over the course of the experiment and believe a 3-10x increase in efficiency is feasible with the exact same approach and without the need for better models.

However, this cost should be put into the perspective that this foundational work of formalization textbook sources can be considered as an infrastructure investment: building foundational libraries that lowers the cost of all future research formalization in that field.

Looking ahead, our results demonstrate that the bottleneck for formalizing the mathematical corpus is no longer a lack of logical reasoning in models but the complex orchestration of their collective labor. This transition from ``hand-crafted'' to ``agent-engineered'' mathematics appears not only possible, but inevitable. In this emerging paradigm, inference compute and capital, rather than human labor and expert knowledge, become the limiting factors for large-scale formalization.
Given the fast trajectory of algorithmic efficiency in machine learning in general \citep{erdil2023algorithmicprogresscomputervision,gundlach2025priceprogressalgorithmicefficiency,modded_nanogpt_2024,epoch2025llminferencepricetrends} and applications in particular \citep[e.g.][]{wu2019accelerating,lin2023evolutionary}, we encourage mathematicians, formalization experts and machine learning scientists to ponder and shape a future of mathematics where formalization provides the backbone of verification and trust in a mathematical research process that scales to the requirements of this century.

\section*{Acknowledgments}
The authors would like to thank Johan Commelin, Bhavik Mehta and other members of the Lean community for feedback and insightful discussions on the project. We thank Darij Grinberg for dedicating his textbooks to the public domain. We thank the FAIR infra teams for technical support.

Machine learning for formal mathematics is indebted to the formal verification community which created the tooling, infrastructure, libraries and datasets that have made large-scale formalization of mathematics with large language models a possibility.

\clearpage
\newpage
\bibliographystyle{assets/plainnat}
\bibliography{paper}

\clearpage
\newpage
\beginappendix

\section{Estimating Token Caching Statistics}
\label{app:cache}

Unfortunately, our logs only contain the total number of input and output tokens per agent dialog, ignoring input caching efficiencies. In this section, we will derive a heuristic estimate for the degree of token caching observed in the experiment and the associated cost savings. Consider the following variables:
\begin{itemize}
\item $N$, number of agents (given),
\item $T$, average number of turns per agent (given),
\item $C$, total number of tokens processed in inputs, with double counting due to reprocessing of previous messages (given),
\item $L$, average total token length of an agent dialog,
\item $m$, average token length per turn.
\end{itemize}
We then have, heuristically (if all dialogs were identical),
\[
C = Nm \sum_{i=1}^T i = Nm \frac{T(T+1)}{2},
\]
or equivalently,
\[
m = \frac{2C}{NT(T+1)}.
\]
Moreover,
\[
L = Tm = \frac{2C}{N(T+1)}.
\]
Assuming prices of
\begin{itemize}
\item $c_{\text{in}}$ for each token processed in inputs,
\item $c_{\text{store}}$ for each token's KV-cache stored,
\item $c_{\text{hit}}$ for each token's KV-cache retrieved,
\end{itemize}
the total input prices $P_{\text{cache}}$ and $P_{\text{nocache}}$ with and without input caching, resp., can be approximated as follows:
\begin{alignat*}{3}
P_{\text{nocache}} &= c_{\text{in}} C
&\quad &= Nmc_{\text{in}}\left(0.5T^2 + 0.5T\right), \\
P_{\text{cache}}   &= c_{\text{hit}} C + (c_{\text{in}} + c_{\text{store}}) NL
&\quad &= Nm c_{\text{in}}\left(0.05T^2 + 3.05T\right),
\end{alignat*}
assuming the current pricing ratios\footnote{\url{https://platform.claude.com/docs/en/build-with-claude/prompt-caching}} $c_{\text{store}} = 2c_{\text{in}}$ for 1 hour and $c_{\text{hit}} = \frac{1}{10} c_{\text{in}}$.

In other words, input caching reduces the coefficient in $T^2$ at the expense of a larger linear coefficient $T$, which is worthwhile for long dialogs like ours (avg. $T \approx 54.8$).

\end{document}

%% file: template/package.tex
\usepackage[T1]{fontenc}
\usepackage[utf8]{inputenc}
\usepackage[english]{babel}

\PassOptionsToPackage{hyphens}{url}
\usepackage{url}
\usepackage{hyperref}

\usepackage{graphicx}
\usepackage{caption}
\usepackage{tikz}
\usepackage{pgfplots}
\pgfplotsset{compat=1.17}

\usepackage{array}
\usepackage{booktabs}
\usepackage{microtype}
\usepackage{xcolor}
\usepackage{enumitem}
\setitemize{itemsep=0pt,parsep=0pt,topsep=0pt}
\setenumerate{itemsep=0pt,parsep=0pt,topsep=0pt}

\usepackage{algorithm}
\usepackage{algorithmicx}
\usepackage{algpseudocode}

\usepackage{natbib}

\usepackage[normalem]{ulem}

%% file: figures/churn_decls.tex
\begin{figure}[b!]
  \centering
  \begin{subfigure}[t]{0.48\linewidth}
    \centering
    \includegraphics[width=\linewidth]{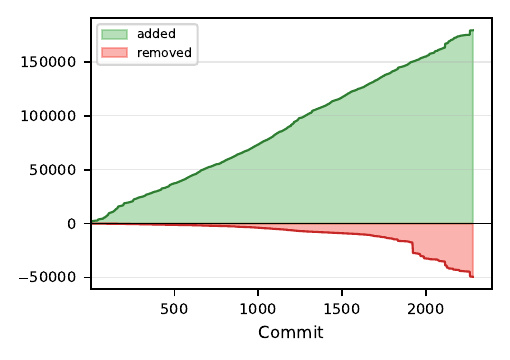}
    \subcaption{\small Added and removed lines}
    \label{fig:churn}
  \end{subfigure}\hfill
  \begin{subfigure}[t]{0.48\linewidth}
    \centering
    \includegraphics[width=\linewidth]{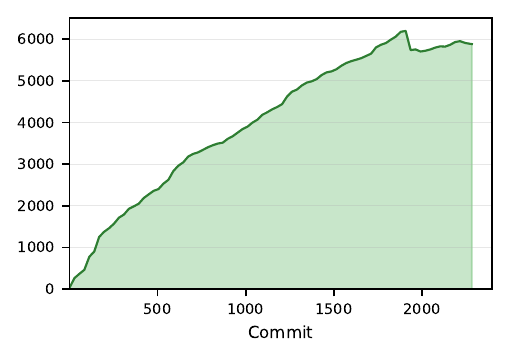}
    \subcaption{\small Lean declarations}
    \label{fig:decls}
  \end{subfigure}
  \caption{\textbf{Progress of the automatic formalization system.} Two phases are clearly discernible: continuous code base expansion (including unnecessary \emph{rabbit hole} content) and cleanup triggered by an update in agent prompts. Each commit represents a separate agent that merges its work into the main branch. At the end, all 340 \emph{target theorems} and definitions are present and proved.}
  \label{fig:churn-decls}
\end{figure}

%% file: figures/summary.tex
\begin{figure}[t]
\centering
\begin{minipage}[b]{0.48\columnwidth}
  \centering
  \vspace{0pt} %
  \begin{tabular}{@{}l r@{}}
    \toprule
    Target theorems and defs. & 340 \\
    \quad of which proved & 340 \\
    Lean source files & 52 \\
    Lines of Lean code & $\sim 130{,}000$ \\
    Lean declarations & $\sim 5{,}900$ \\
    Successful agents & $\sim 3{,}300$ \\
    Pages in source text & $\sim 500$ \\
    \bottomrule
  \end{tabular}
  \vspace{2em}
  \captionof{table}{Formalization statistics}
  \label{tab:stats}
\end{minipage}\hfill
\begin{minipage}[b]{0.4\columnwidth}
  \centering
  \vspace{0pt} %
  \includegraphics[width=\linewidth]{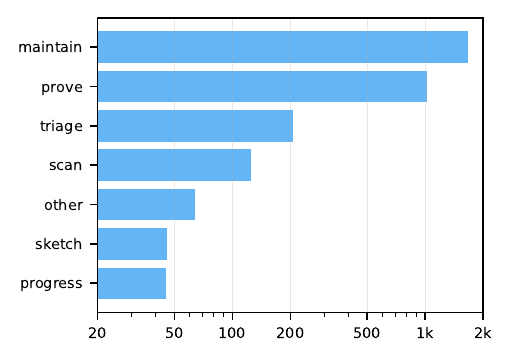}
  \caption{Number of agents by type}
  \label{fig:agents-by-type}
\end{minipage}
\end{figure}

%% file: figures/contents.tex
\begin{figure}[b!]
\centering
\begin{tcolorbox}[
    colback=myboxcolor, %
    colframe=blue!40!black, %
    boxrule=1pt, %
    arc=4pt, %
    width=0.85\textwidth,
    title={Content overview},
    fonttitle=\bfseries
]
{\small
\textbf{Formal power series:} limits, division, derivatives, substitution, exponentials and logarithms, non-integer powers,  generalized binomial theorem, Laurent series, multivariate series, infinite products, multipliability, distributive laws.

\textbf{Permutations:} sign, cycle decomposition, inversions and Lehmer codes, length generating function.

\textbf{Determinants:} Cauchy--Binet formula, Desnanot--Jacobi identity, Dodgson condensation, Jacobi's complementary minor theorem, Vandermonde determinant.

\textbf{Alternating sums and signed counting:} alternating-sum cancellations, sign-reversing involutions, inclusion--exclusion, weighted inclusion--exclusion, Boolean Möbius inversion, derangements, surjection counting.

\textbf{Partitions and q-series:} partitions and Ferrers diagrams, conjugation, partition generating function, odd parts and distinct parts correspondence, Euler pentagonal number theorem, Jacobi triple product identity, divisor-sum recurrences.

\textbf{q-binomial coefficients:} q-integers, q-factorials, Gaussian binomial coefficients, q-binomial theorems, subspace-counting interpretation, classical limits.

\textbf{Symmetric functions:} symmetric polynomials, elementary complete homogeneous and power-sum families, Newton--Girard relations, monomial symmetric polynomials and basis theorem, Schur polynomials, skew Schur polynomials, alternants, Bender--Knuth involutions, Pieri rules, Jacobi--Trudi identities, Littlewood--Richardson rule.

\textbf{Lattice paths:} lattice path enumeration, Lindström--Gessel--Viennot lemma, weighted path model, nonintersecting path families.

\textbf{Applications:} domino tilings (height-2 Fibonacci recurrence, height-3 classification).
}
\end{tcolorbox}
\caption{Overview of the contents from the source textbook \citep{grinberg2025introduction}.}
\label{fig:contents}
\end{figure}

%% file: figures/loc_hists.tex
\begin{figure}[t]
  \centering
  \begin{subfigure}[t]{0.32\linewidth}
    \centering
    \includegraphics[width=\linewidth]{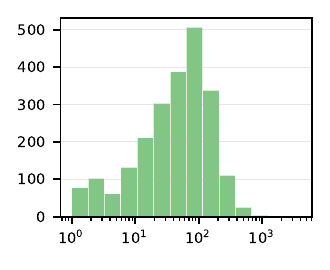}
    \subcaption{\small Added lines}
    \label{fig:hist-add}
  \end{subfigure}\hfill
  \begin{subfigure}[t]{0.32\linewidth}
    \centering
    \includegraphics[width=\linewidth]{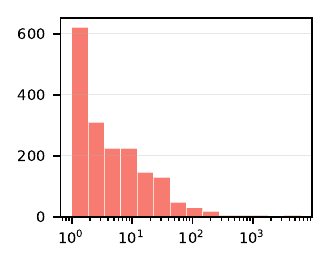}
    \subcaption{\small Removed lines}
    \label{fig:hist-del}
  \end{subfigure}\hfill
  \begin{subfigure}[t]{0.32\linewidth}
    \centering
    \includegraphics[width=\linewidth]{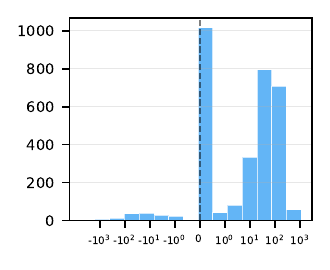}
    \subcaption{\small Net change}
    \label{fig:hist-net}
  \end{subfigure}
  \caption{\textbf{Histograms of agent PR line counts.}
  The histograms show the number of agents that return within a given range of added, removed or net added lines. 
  The multi-agent setup ensures that pull requests are granular and atomic, comprising around 100 lines.
  Refactorings that reduce the overall code size are rare, but PRs frequently replace code while extending the code base overall.
  }
  \label{fig:hists}
\end{figure}

%% file: figures/agent_tokens.tex
\begin{table}[tbp]
\centering
\begin{tabular}{lrrrrrrrr}
\toprule
\textbf{Agent Type} & \textbf{Count} & \textbf{In (M)} & \textbf{Out (M)} & \textbf{Total (M)} & \textbf{Avg In (K)} & \textbf{Avg Out (K)} & \textbf{Turns} & \textbf{Avg Turns} \\
\midrule
Sketcher & 85 & 6 & 0.0 & 6 & 71 & 0.4 & 5084 & 59.8 \\
Prover & 8704 & 25012 & 194.2 & 25206 & 2874 & 22.3 & 435471 & 50.0 \\
Maintainer & 6467 & 44770 & 277.1 & 45047 & 6923 & 42.9 & 814363 & 125.9 \\
Math Reviewer & 6797 & 3759 & 42.9 & 3802 & 553 & 6.3 & 147753 & 21.7 \\
Eng. Reviewer & 6805 & 1542 & 20.7 & 1563 & 227 & 3.0 & 86118 & 12.7 \\
Triage & 550 & 2747 & 13.1 & 2760 & 4994 & 23.8 & 68827 & 125.1 \\
Scan & 307 & 4395 & 9.6 & 4405 & 14317 & 31.3 & 72170 & 235.1 \\
Progress & 331 & 944 & 3.4 & 948 & 2853 & 10.4 & 15488 & 46.8 \\
\midrule
\textbf{Total} & 30046 & 83176 & 561.2 & 83737 & 2768 & 18.7 & 1645274 & 54.8 \\
\bottomrule
\end{tabular}
\caption{\textbf{Token usage by agent type.} See Sections~\ref{sec:dev-agents}~and~\ref{sec:coord} for descriptions of the different roles.}
\label{tab:agent-tokens}
\end{table}

%% file: figures/outcome_tokens.tex
\begin{table}[tbp]
\centering
\begin{tabular}{lrrrrrrrr}
\toprule
\textbf{Outcome} & \textbf{Count} & \textbf{In (M)} & \textbf{Out (M)} & \textbf{Total (M)} & \textbf{Avg In (K)} & \textbf{Avg Out (K)} & \textbf{Turns} & \textbf{Avg Turns} \\
\midrule
Merged & 3490 & 16477 & 96.4 & 16574 & 4721 & 27.6 & 299143 & 85.7 \\
Approved & 589 & 1298 & 7.2 & 1305 & 2203 & 12.2 & 20832 & 35.4 \\
Max Revisions & 976 & 6935 & 46.5 & 6981 & 7105 & 47.7 & 131609 & 134.8 \\
Max Iterations & 11 & 101 & 0.8 & 102 & 9216 & 70.4 & 1408 & 128.0 \\
No PR & 14192 & 7448 & 76.4 & 7524 & 525 & 5.4 & 259003 & 18.2 \\
No PR (blocked) & 4668 & 8012 & 55.8 & 8068 & 1716 & 12.0 & 146026 & 31.3 \\
Aborted & 6120 & 42904 & 278.0 & 43182 & 7011 & 45.4 & 787253 & 128.6 \\
\midrule
\textbf{Total} & 29691 & 80686 & 542.7 & 81228 & 2718 & 18.3 & 1605312 & 54.1 \\  %
\bottomrule
\end{tabular}
\caption{\textbf{Token usage by agent outcome.} 
See Section~\ref{sec:outcomes} for the taxonomy of agent outcomes.
}
\label{tab:outcome-tokens}
\end{table}

%% file: figures/outcomes_runtime.tex
\begin{figure}[tb]
  \centering
  \includegraphics[width=\linewidth]{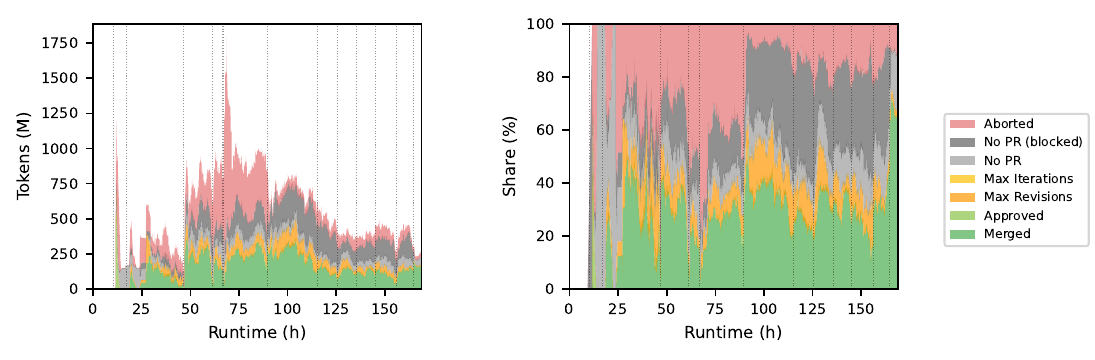}
  \caption{\textbf{Agent outcomes over time by amount (left) and proportion (right) of tokens spent.}
  See Section~\ref{sec:outcomes} for the taxonomy of agent outcomes.
  The data is averaged over rolling windows of 400 agents and excludes short debugging runs that marginally contributed to the repository. Dotted vertical lines indicate interventions or restarts (typically minor code changes for efficiency and better observability). Detailed token statistics are not available for the first hours of the run.
  It can be seen that the initial code suffered from performance issues (especially regarding NFS bottlenecks, \texttt{git worktree} timeouts and merge queue congestion), resulting in a large number of tokens spent on \emph{aborted} agents.
  Later versions removed these problems, and mainly suffered from issue dependencies as evidenced by a large number of \emph{blocked} agents. This could be resolved with a targeted intervention of \emph{status agents} that cleaned up stale issues, analyzed the dependencies and suggested a course of action, resulting in an increased success rate in the latest stage of the run.
  Note that merged PRs can also exclusively touch issue files, especially in the case of \emph{maintainer} agents, and hence success rates of close to 100\% should be achievable.
  }
  \label{fig:outcomes-runtime}
\end{figure}

%% file: figures/prover_maintainer.tex
\begin{figure}[t]
  \centering
  \begin{subfigure}[t]{0.4\linewidth}
    \centering
    \includegraphics[width=\linewidth]{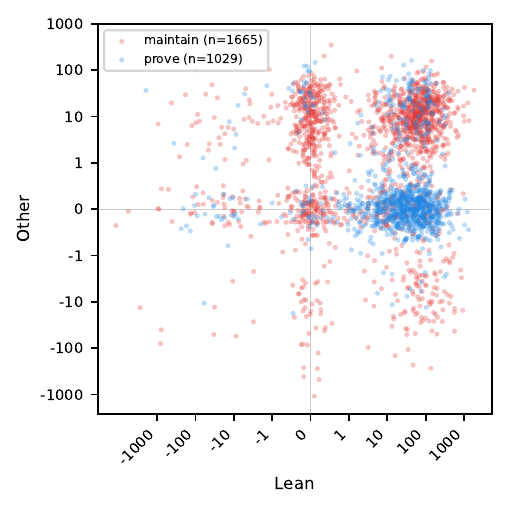}
    \subcaption{\small Net line change statistics}
    \label{fig:prover-maintainer-scatter}
  \end{subfigure}\hfill
  \begin{subfigure}[t]{0.48\linewidth}
    \centering
    \raisebox{4ex}{\includegraphics[width=\linewidth]{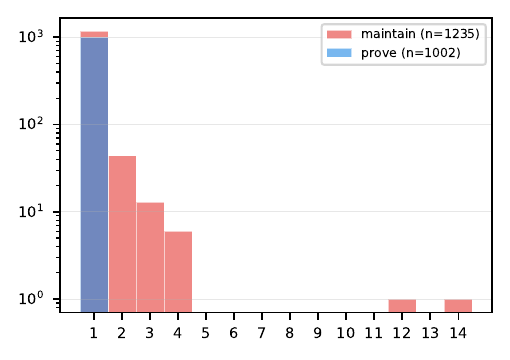}}
    \subcaption{\small Histogram of the number of Lean files edited}
    \label{fig:prover-maintainer-hist}
  \end{subfigure}
  \caption{\textbf{Comparing prover and maintainer agents.}
    \textbf{ (a)} Net line change statistics on Lean files and coordination files (``other'') for \emph{prover} and \emph{maintainer} agents. For readability, the plot uses a symlog axis and Gaussian jitter with $\sigma = 0.3$ grid units.
    \textbf{(b)} Histogram of the number of Lean files edited for both agent types. \emph{Prover} agents work on a single file by design, \emph{maintainer} agents most frequently, too, but can also conduct large refactorings involving up to 14 files.  
  }
  \label{fig:prover-maintainer}
\end{figure}